\documentclass[sigconf]{acmart}
\acmSubmissionID{549}

\usepackage{booktabs} 
\usepackage{amsmath}
\usepackage{colortbl}
\usepackage[ruled]{algorithm2e} 

\SetAlFnt{\small}
\SetAlCapFnt{\small}
\SetAlCapNameFnt{\small}
\SetAlCapHSkip{0pt}

\def\MethodName{Dr.3D}

\copyrightyear{2022}
\acmYear{2022}
\setcopyright{acmcopyright}
\acmConference[SA '22 Conference Papers]{SIGGRAPH Asia 2022 Conference Papers}{December 6--9, 2022}{Daegu, Republic of Korea}
\acmBooktitle{SIGGRAPH Asia 2022 Conference Papers (SA '22 Conference Papers), December 6--9, 2022, Daegu, Republic of Korea}
\acmPrice{15.00}
\acmDOI{10.1145/3550469.3555422}
\acmISBN{978-1-4503-9470-3/22/12}

\begin{document}
\title{\MethodName{}: Adapting 3D GANs to Artistic Drawings}

\settopmatter{authorsperrow=3}
\author{Wonjoon Jin}
\orcid{0000-0001-6883-3920}
\affiliation{%
  \institution{POSTECH}
   \country{South Korea}
}
\email{jinwj1996@postech.ac.kr}

\author{Nuri Ryu}
\orcid{0000-0002-7769-689X}
\affiliation{%
  \institution{POSTECH}
   \country{South Korea}
}
\email{ryunuri@postech.ac.kr}

\author{Geonung Kim}
\orcid{0000-0003-0806-6963}
\affiliation{%
  \institution{POSTECH}
   \country{South Korea}
}
\email{k2woong92@postech.ac.kr}

\author{Seung-Hwan Baek}
\orcid{0000-0002-2784-4241}
\affiliation{%
  \institution{POSTECH}
   \country{South Korea}
}
\email{shwbaek@postech.ac.kr}

\author{Sunghyun Cho}
\orcid{0000-0001-7627-3513}
\affiliation{%
  \institution{POSTECH}
   \country{South Korea}
}
\affiliation{%
  \institution{Pebblous}
   \country{South Korea}
}
\email{s.cho@postech.ac.kr}

\renewcommand{\shortauthors}{Jin et al.}



\begin{teaserfigure}
    \centering
    \includegraphics[width=0.98\linewidth]{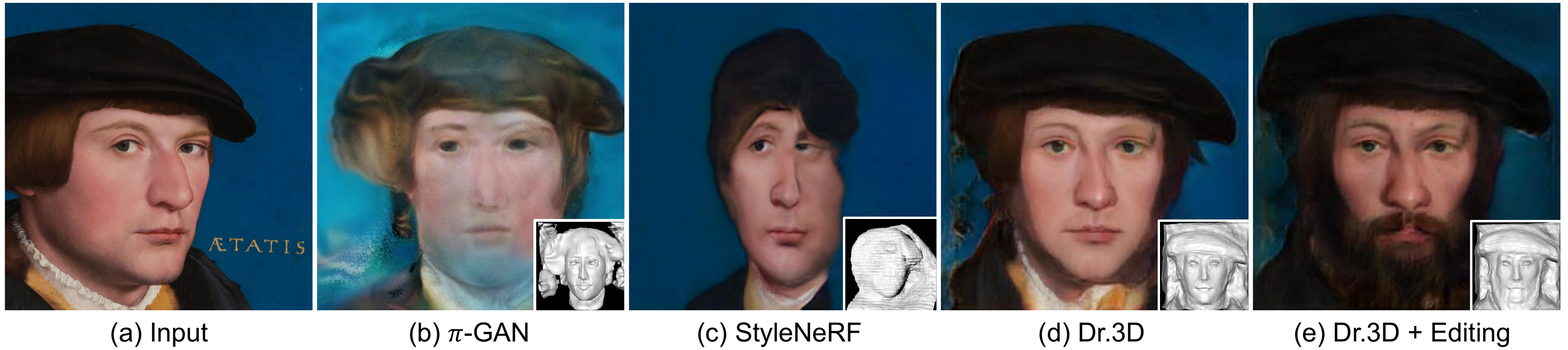}
        \caption{
        GAN inversion and semantic editing examples on a portrait drawing.
        For comparison, we perform na\"ive domain adaptation to $\pi$-GAN~\cite{pigan} and StyleNeRF~\cite{stylenerf} by finetuning them on portrait drawings.
        Then, we invert the input image in (a) using an off-the-shelf GAN inversion method to a latent code and reconstruct the image and its shape at a different camera pose using each 3D GAN model.
        The results in (b) and (c) show that na\"ive adaptations of existing 3D GANs fail to handle the input drawing.
        On the other hand, our method can successfully reconstruct the input image, and also allow semantic editing as shown in (d) and (e). 
Image in (a): Portrait of a Member of the Wedigh Family, 1532 by Hans Holbein the Younger, WikiArt [Public Domain] via (https://bit.ly/3KfgKPI)
        }
    \label{fig:teaser}
\end{teaserfigure}


\begin{abstract}
While 3D GANs have recently demonstrated the high-quality synthesis of multi-view consistent images and 3D shapes, they are mainly restricted to photo-realistic human portraits.
This paper aims to extend 3D GANs to a different, but meaningful visual form: artistic portrait drawings.
However, extending existing 3D GANs to drawings is challenging due to the inevitable geometric ambiguity present in drawings.
To tackle this, we present \MethodName{}, a novel adaptation approach that adapts an existing 3D GAN to artistic drawings.
\MethodName{} is equipped with three novel components to handle the geometric ambiguity: a deformation-aware 3D synthesis network, an alternating adaptation of pose estimation and image synthesis, and geometric priors.
Experiments show that our approach can successfully adapt 3D GANs to drawings and enable multi-view consistent semantic editing of drawings.
\end{abstract}

%
%


%
%


\maketitle

\section{Introduction}
\label{sec:intro}

Generative adversarial networks (GANs)~\cite{gan} have achieved remarkable success in learning to synthesize realistic images, which is crucial for a plethora of applications in computer graphics and vision~\cite{stylegan, stylegan2}.
Notably, GANs allow us to explore synthesized images and edit real images in a semantically meaningful way~\cite{interfacegan, sefa, stylespace}.
Among many image categories that GAN methods have dealt with, it is not surprising that the human face is one of the most popular targets in computer graphics and vision.
Recently, making GANs aware of 3D geometry has received great attention, opening up an exciting research field of 3D GANs.
They tackle the ill-posed problem of learning the 3D-aware distribution of real images by explicitly modeling 3D light transport between a camera and a target object.
3D GANs enable the synthesis and editing of photographs not only in a semantically meaningful way, but also in consideration of 3D scene geometry~\cite{eg3d, pigan, stylenerf, cips, giraffe}.

To date, 3D GANs have been mainly demonstrated only on real-world photographs, which are the exact recordings of real-world scenes through perspective cameras. 
In this paper, we aim to extend the capability of 3D GANs to handle a different, but meaningful visual form: \emph{drawing}.
Drawing plays a crucial role in human history by depicting both real-world and imaginary subjects with intended and/or unintended variations.
Existing 2D GAN methods have been extended to cope with drawings by adapting 2D GANs pretrained on real-world photographs into drawings, so-called domain adaptation~\cite{stylegan-ada, fewshot-correspondence, cyclegan, image2image}.
The adaptation strategy exploits common features between photographs and drawings, allowing us to bring the synthesis and editing capability of 2D GANs to the drawing domain~\cite{stylealign}. 
Unfortunately, extending 3D GANs to the drawing domain turns out to be more challenging as shown in Figure~\ref{fig:teaser}.  

One fundamental reason for this difficulty is that drawings have intrinsic geometric ambiguity on the subject and camera pose. 
Artists intentionally or unintentionally assume nondeterministic geometry of subjects from an imaginary viewpoint deviating from the physical one, resulting in drawing with creative ambiguity.
This further increases the ill-posedness of learning a 3D-aware image distribution of drawings
and hinders the direct application of previous domain adaptation methods used in 2D GANs for 3D GAN methods.
Figure~\ref{fig:teaser} shows that the application of state-of-the-art 3D GANs~\cite{stylenerf, pigan} on drawings via domain adaptation fails to synthesize faithful 3D-consistent images.

This paper proposes \MethodName{}, a novel 3D GAN domain adaptation method for portrait drawings.
\MethodName{} effectively handles the fundamental geometric ambiguity of drawings with three remedies.
First, we present a deformation-aware 3D synthesis network suitable for learning a large distribution of diverse shapes in drawing.
Second, we propose an alternating adaptation scheme for 3D-aware image synthesis and pose estimation, effectively reducing the learning complexity of ambiguous 3D geometries and camera poses in drawings.
Third, we impose geometric priors that enable stable domain adaptation from real photographs to drawings.
The resulting domain adaptation method, \MethodName{}, is the first method that enables stable editing and synthesis of drawing images in a 3D consistent way.
We validate the effectiveness of \MethodName{} via extensive quantitative and qualitative evaluations.

\section{Related works}

\paragraph{3D-aware GANs}
Several recent works have extended 2D GANs to be aware of the 3D structures of subjects and camera poses. 
Voxel-based 3D GANs~\cite{hologan} directly represent 3D structures with 3D voxel grids parameterized by 3D convolutional neural networks.
Unfortunately, they typically suffer from large memory requirements. 
Mesh-based GANs~\cite{mesh3dgan1, mesh3dgan2} lift the memory problem by using sparse meshes as a geometric representation.
However, dealing with such sparse primitives with neural networks is challenging due to their unstructured data types.
Recently, implicit 3D GANs~\cite{eg3d, pigan, stylenerf, giraffe, graf} have shown promising performance in terms of image fidelity and 3D consistency.
GRAF and $\pi$-GAN~\cite{graf, pigan} first proposed to learn to generate neural radiance fields (NeRF)~\cite{nerf} and synthesize images via differentiable volumetric rendering.
Since then, several attempts have been made to further improve the synthesis quality by incorporating feature projection and upsampling with the expense of losing multi-view consistency~\cite{giraffe, stylenerf}.
Most recently, EG3D~\cite{eg3d} demonstrates the synthesis of high-resolution 3D-aware images based on a tri-plane representation and a StyleGAN generator~\cite{stylegan2}.  
Albeit great progress has been made in 3D GANs, directly applying them to drawings fails to learn meaningful 3D structures due to the large domain gap between real photographs and drawings (Figure~\ref{fig:teaser}).

\paragraph{Photo-to-Drawing Domain Adaptation}
Applying GANs to drawings has often been practiced via domain adaptation in the 2D image space where we first train a GAN model on real photographs, and then finetune the model on a drawing dataset.
This domain-adaptation technique has achieved notable success in synthesizing high-fidelity drawing images.
Moreover, the adapted models inherit the semantically-meaningful editing capability of previous 2D GANs, thus enabling semantic editing of drawing images.
Thus, their applications span the diverse computer graphics and vision fields, resulting in new applications such as image cartoonization~\cite{toonify, dualstylegan} and automatic caricature generation~\cite{stylecarigan}. 

However, extending the success of 2D GANs to 3D GANs has been challenging.
Drawings have ambiguous and diverse geometric shapes and appearances, resulting in a large domain gap between real photographs and drawings as witnessed by recent works~\cite{stylenerf}.
Typical failure examples are flattened 3D shapes, inconsistent multi-view images, and low-fidelity images as shown in Figure~\ref{fig:qualitative_comparison}.
We aim to overcome this hurdle by proposing a 3D domain adaptation method designed explicitly for drawings and demonstrates compelling results via our stable photo-to-drawing domain adaptation.

\paragraph{Non-generative 3D-aware Image Editing}
Editing an input image considering its 3D structure can be also done without using generative models.
For instance, fitting a 3D parametric shape model~\cite{3dmm, FLAME, bfm} to an image allows us to have a geometrically-editable 3D model textured with the image~\cite{deep3dface, DECA}. 
StyleRig~\cite{stylerig} propose combining 3DMM parameters with semantic features learned in 2D GANs for image editing.
Unfortunately, parametric shape models are not applicable to drawings as the diverse 3D geometries in drawings often deviate from the representation space of existing parametric shape models. 
Another research direction is to reconstruct the 3D geometry of an input image based on the physics-based priors of light transport, where Unsup3d~\cite{unsup3d}, GAN2Shape~\cite{gan2shape}, and StyleGANRender~\cite{imagegansmeet} show promising results. 
However, these methods often fail to handle drawings, because of their restrictive physics-based priors that assume the accurate decomposition of an image into illumination, appearance, and shape, which does not hold in drawings.

\section{Background on EG3D}
Before introducing our approach, we first provide a brief review of the network architecture of EG3D~\cite{eg3d}, a state-of-the-art 3D GAN network upon which our network is built.
Specifically, it starts with a randomly sampled GAN latent code $z$, which turns into 2D convolutional features after passing through a StyleGAN-based feature generator~\cite{stylegan}. 
Generated 2D feature maps are then rearranged into 3D orthogonal feature planes, from which any 3D point can be described with the projected features. 
Given the features, a multi-layer perceptron (MLP) decoder predicts the color and density of a 3D point, which are subsequently used for volume rendering, resulting in an image ${x}_\mathrm{fake}$ and a depth map ${d}_{fake}$ at a camera pose $\theta$.
The feature generator and the MLP decoder are trained using a discriminator $D$, which is conditioned with the input camera pose $\theta$ to promote the generator to synthesize images that accurately reflect the camera pose.
In contrast to being successful as a 3D GAN model for realistic portrait images, directly applying EG3D to drawings results in catastrophic failures as shown in Figure ~\ref{fig:ablation_img}, due to the fundamental ambiguity in drawings.
\section{Domain Adaptation to Drawings}

Built upon EG3D~\cite{eg3d}, \MethodName{} is equipped with three remedies that mitigate the ill-posedness of photo-to-drawing 3D-aware domain adaptation: (1) a deformation-aware 3D synthesis network, (2) an alternating adaptation scheme for image synthesis and pose estimation, and (3) geometric priors for adaptation to drawing.
In this section, we introduce each remedy in detail.

\begin{figure}[!t]
\centering
\includegraphics[width=0.95\linewidth]{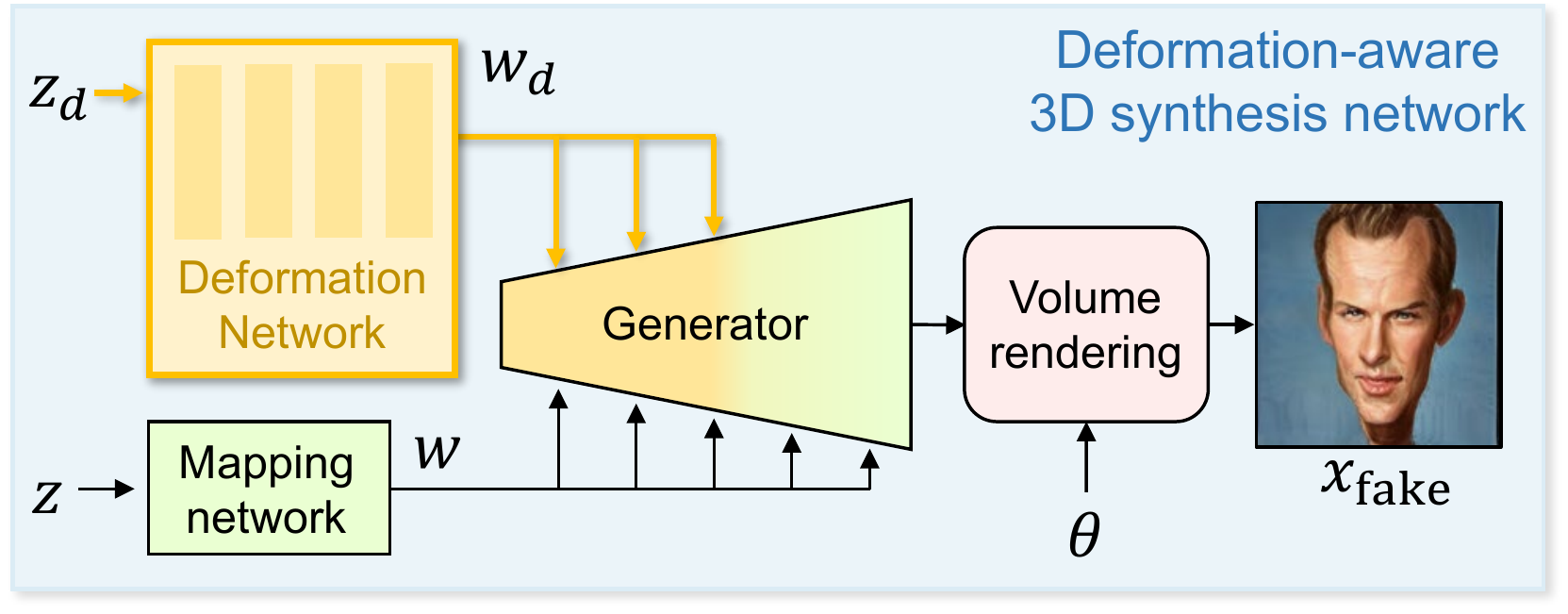}
\vspace{-0.2cm}
\caption{
Network architecture of a deformation-aware 3D synthesis network.
The network consists of a deformation network, a mapping network, a feature generator, and a volume rendering module.
The network takes latent codes $z_d$ and $z$, and a camera pose parameter $\theta$ as inputs, and synthesizes an image in a multi-view consistent way.
}
\label{fig:framework}
\end{figure}

\begin{figure}[t]
\centering
\includegraphics[width=0.95\linewidth]{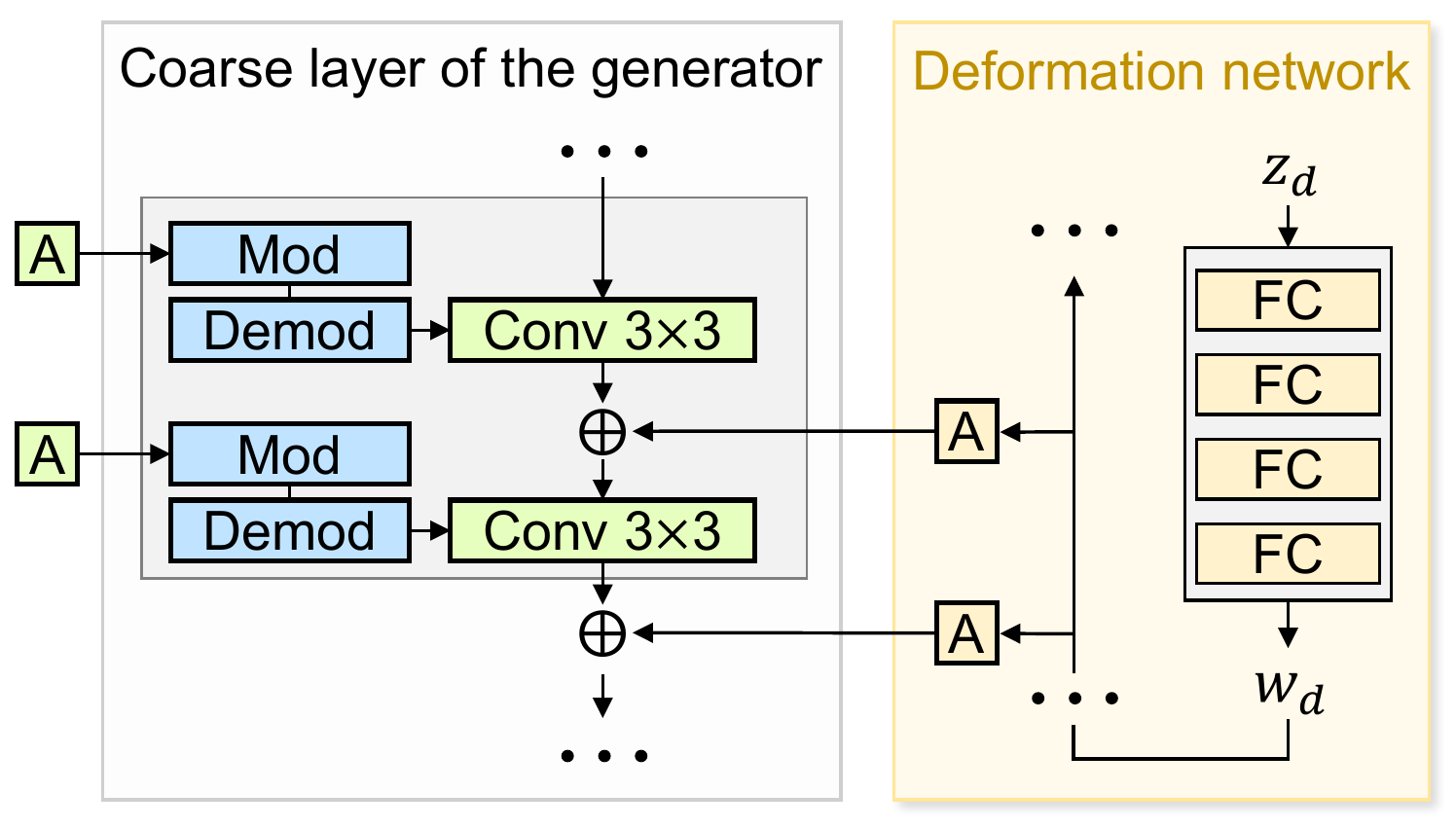}
\vspace{-0.2cm}
\caption{
Network architectures of a generator and a deformation network.
The generator network is based on the StyleGAN2 generator~\cite{stylegan2}.
FC: a fully-connected (FC) layer. A: an affine layer consisting of a single FC layer. Mod: a modulation layer. Demod: a demodulation layer.
}
\label{fig:deform_network}
\end{figure}

\subsection{Deformation-aware 3D Synthesis}
\label{sec:deform_net}
Drawings may have local shape variations that do not exist in photographs taken by cameras.
To handle such domain gaps effectively, we introduce a deformation-aware 3D synthesis network $G$.
Our network architecture builds on top of the EG3D network as shown in Figure~\ref{fig:framework}.
To model diverse shape deformations in drawings, our network uses an additional latent code $z_d$.
The deformation code $z_d$ turns into residual features via an MLP-based deformation network as shown in Figure~\ref{fig:deform_network}, which are then added to early convolutional features in the StyleGAN feature generator.
Note that modulating early layers in a StyleGAN generator is known to provide large-scale changes to synthesized images~\cite{stylecarigan, dualstylegan}.
This simple feature-modulation strategy allows us to model diverse shape variations in drawings.

The role of the deformation-aware network is twofold.
First, it introduces additional dimensions to the latent space so that local shape variations that may uniquely exist in target artistic drawing domains can be more effectively handled.
Second, the residual features generated by the deformation-aware network help model the domain gap between the source and target domains more effectively. Specifically, as the mapping network is an MLP and the generator consists of spatially-invariant convolution operations, finetuning them cannot effectively model local deformations. To resolve this, our deformation-aware network estimates spatially-variant residual features to better handle local feature differences between the source and target domains. Moreover, the residual features help retain the original weights of the generator so that the knowledge about 3D structures learned in the original networks can be better preserved for more successful domain adaptation.
Refer to the Supplemental Document for implementation details and additional analysis of the deformation network.

\subsection{Alternating Adaptation of Pose Estimation and Image Synthesis}
\label{sec:adapt_pose}
EG3D~\cite{eg3d}, which our method builds upon, requires known camera poses associated with input training images for its training.
While camera poses for real portraits can be readily estimated using an off-the-shelf pose estimation network~\cite{deep3dface,DECA},
it is not trivial to obtain camera poses for portrait drawings.
Previous pose estimation networks trained on real portraits fail on drawings due to the large domain gap,
and there exist no datasets with ground-truth poses of drawings to train a pose estimation network.
To tackle this problem, we may also adapt a pose-estimation network trained on portrait photos to drawings so that we can estimate the poses of drawings to train a synthesis network.
However, adapting a pose-estimation network and a 3D synthesis network is a chicken-and-egg problem.
Adapting a pose-estimation network requires training data with ground-truth pose labels, which can be obtained by an adapted 3D synthesis network,
while adapting a 3D synthesis network requires an accurately adapted pose-estimation network.
\begin{figure*}[!t]
\centering
\includegraphics[width=0.98\linewidth]{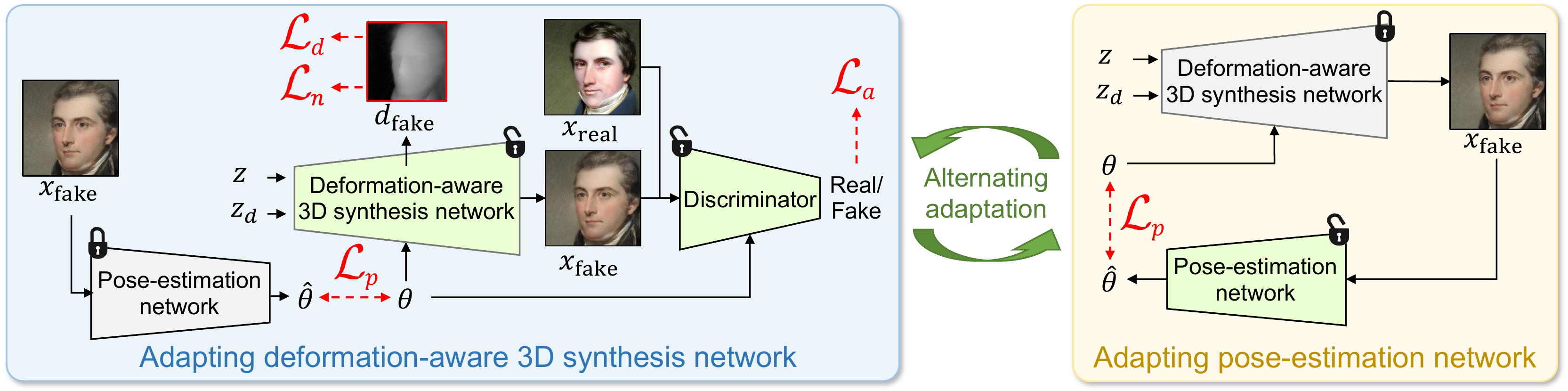}
\vspace{-0.2cm}
\caption{
Alternating adaptation. Our approach alternatingly adapts the deformation-aware 3D synthesis network and the pose-estimation network. 
$x_\mathrm{real}$: Portrait of Benjamin Moore McVickar, 1825 by Charles Cromwell Ingham, MetMuseum [Public Domain] via (https://bit.ly/3c9skiy).
}
\label{fig:alternate_optim}
\end{figure*}

To resolve this, we propose an alternating adaptation approach that alternatingly updates the 3D synthesis network $G$ and pose-estimation network $P$ (Figure~\ref{fig:alternate_optim}).
Specifically, at each iteration of the alternating adaptation,
we synthesize a pseudo ground-truth dataset using the current $G$, and update $P$ using the synthesized dataset.
Then, using the updated $P$, we estimate the poses of the real drawings in a training dataset and update $G$ using the estimated poses.
In this way, we can progressively adapt both $P$ and $G$ to a target drawing domain.
However, at early iterations of the alternating adaptation, the poses of training images are not accurately estimated by $P$ due to the large domain gap, which may eventually lead to the failure of adaptation.
To overcome this, we introduce training losses with geometric priors, which will be described in Section \ref{sec:geo_prior}, to guide the adaptation process.
In the following, we describe each step of our alternating adaptation in more detail.


\paragraph{Adapting 3D Synthesis Network}
Given an input drawing $x_\mathrm{real}$ as a training sample, we estimate its camera pose $\theta$ using a \emph{fixed} pose estimation network.
With the estimated pose $\theta$, our 3D synthesis network $G$ generates an image $x_\mathrm{fake}$ and its corresponding depth map $d_\mathrm{fake}$.
To adapt $G$, we employ the adversarial loss $\mathcal{L}_{a}$ of the original EG3D~\cite{eg3d}, which is based on a conditional discriminator $D$.
Specifically, $D$ takes either a synthetic or real image, $x_\mathrm{fake}$ or $x_\mathrm{real}$, with its corresponding camera pose $\theta$ and evaluates how realistic it is.
We update both $G$ and $D$ an adversarial-learning manner by back-propagating the loss.
However, using the adversarial loss alone is not enough as there is no guarantee that the camera pose $\theta$ is accurate especially at early iterations of the alternating adaptation.
Inaccurate pose estimation typically leads to learning flattened geometries for drawings as shown in Figure \ref{fig:ablation_img}.
To address this issue, we introduce an additional loss $\mathcal{L}_g$ based on geometric priors, described in Section \ref{sec:geo_prior}.
The 3D synthesis network $G$ is then updated by minimizing a loss defined as:
\begin{equation}
    \mathcal{L} = \mathcal{L}_a (x_\textrm{fake}, x_\textrm{real}, \theta) + \mathcal{L}_g(x_\textrm{fake},d_\textrm{fake},\theta) .
\end{equation}

\paragraph{Adapting Pose-estimation Network.}
We adapt the pose-estimation network $P$ while fixing the 3D synthesis network $G$.
To adapt $P$, we first generate a pseudo training dataset $\Omega$ that consists of multiple pairs of randomly sampled camera poses $\theta$ and their corresponding images $x_\textrm{fake}^\theta$.
We synthesize $x_\textrm{fake}^\theta$ as $x_\textrm{fake}^\theta = G(z,\theta)$ where $z$ is a randomly sampled GAN latent code.
On the pseudo dataset, we finetune our pose-estimation network $P$ by minimizing the pose-estimation loss $\mathcal{L}_p$ defined as:
\begin{equation} \label{eq_posenet}
\begin{aligned}
&\mathcal{L}_{p} = \frac{1}{|\Omega|} \sum_{\{\theta,x_\mathrm{fake}^\theta\} \in \Omega} \left\lVert\theta-P(x_\mathrm{fake}^\theta)\right\rVert^2_2.
\end{aligned}
\end{equation}
As our deformation-aware 3D synthesis network $G$ continuously adapts to a drawing domain thanks to the adversarial and geometric-prior-based losses, our pose-estimation network $P$ can coordinately adapt to a drawing domain through alternating adaptation. 

\subsection{Additional Losses with Geometric Priors}
\label{sec:geo_prior}
In order to guide the alternating adaptation process to a proper solution, the loss $\mathcal{L}_g$ is defined as a combination of three losses:
\begin{equation} \label{eq_prior}
\begin{aligned}
\mathcal{L}_{g} = \alpha \mathcal{L}_{d} + \beta \mathcal{L}_{n} + \gamma \mathcal{L}_{p},
\end{aligned}
\end{equation}
where $\alpha$, $\beta$ and $\gamma$ are balancing weights.
$\mathcal{L}_{d}$ is a depth similarity loss, $\mathcal{L}_{n}$ is a normal smoothness loss, and $\mathcal{L}_{p}$ is a pose loss defined in Equation~\eqref{eq_posenet}.
The pose loss $\mathcal{L}_{p}$ guides the 3D synthesis network to synthesize an image that matches the input camera pose $\theta$.
$\mathcal{L}_d$ and $\mathcal{L}_n$ correspond to geometric priors that guide $G$ to synthesize a valid 3D geometry and an image correctly reflecting the input camera pose $\theta$.
In the following, we describe geometric priors $\mathcal{L}_d$ and $\mathcal{L}_n$ in detail, and discuss how the loss terms guide the alternating adaptation process to a proper solution.

\paragraph{Depth Similarity Loss}
Even though portrait drawings have intrinsic geometric ambiguity, there are still similarities between drawings and real photographs because the category of subjects is still the same as human face. 
This incurs our first observation: the geometry of a subject depicted in a drawing is similar to the geometry in a photograph \emph{at a high level}. We implement such prior by penalizing the \emph{low-frequency} difference between the depth of a synthesized drawing $d_\mathrm{fake}$ and that of a synthesized photo $d_\mathrm{fake, photo}$: 
\begin{equation} \label{eq_depth}
\begin{aligned}
\mathcal{L}_{d} = \left\lVert k*d_\mathrm{fake}-k*d_\mathrm{fake, photo}\right\rVert^2_2,
\end{aligned}
\end{equation}
where $k$ is a $15\times15$-sized Gaussian low-pass filter of standard deviation 5.
We use a synthesis network $G_\mathrm{photo}$ trained on real FFHQ photos~\cite{stylegan} to generate its depth $d_\mathrm{fake, photo}=G_\mathrm{photo}(z,\theta)$.
Note that latent code $z$ and pose $\theta$ are the ones used for the drawing sample:  $d_\mathrm{fake}=G(z,\theta)$.
\begin{figure*}[!t]
\centering
\includegraphics[width=0.95\linewidth]{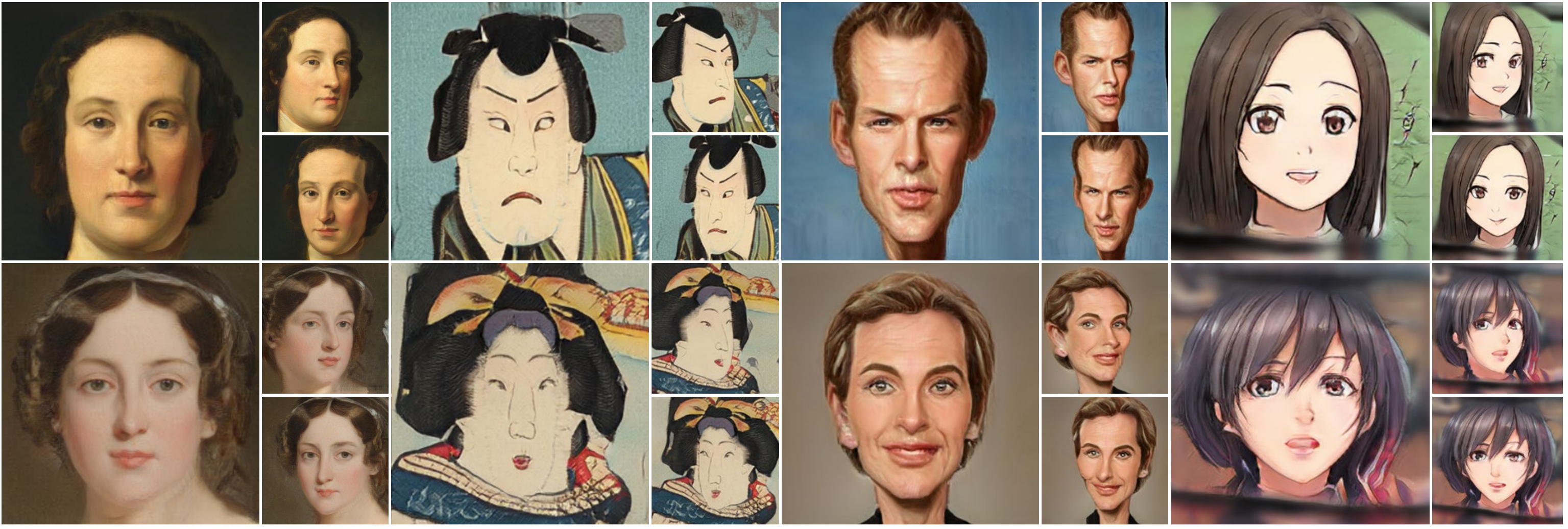}
\vspace{-0.2cm}
\caption{3D-aware drawing synthesis results of our 3D synthesis network adapted to different datasets by \MethodName{} (from left to right: historical art, ukiyo-e, caricature, and anime).
}
\label{fig:curated_examples}
\end{figure*}

\paragraph{Normal Smoothness Loss}
We further penalize abrupt changes of a synthesized geometry, which is implemented as a loss function:
\begin{equation} \label{eq_smoothloss}
\mathcal{L}_{n} = \| \nabla n_\mathrm{fake} \|_2^2, 
\end{equation}
where $\nabla$ is the spatial gradient operator and $n_\mathrm{fake}$ is a surface normal map computed from a synthesized depth map $d_\mathrm{fake}$.

\paragraph{Effect on Alternating Adaption}
The additional losses are crucial in guiding the alternating adaptation toward a proper solution.
At early iterations of the alternating adaptation process, the 3D synthesis network $G$ produces images that are close to real portrait images.
As the pose estimation network $P$ can accurately estimate the camera poses of such synthesized images at early iterations, the pose loss $\mathcal{L}_p$ can enforce $G$ to produce images of the correct camera poses.
On the other hand, the depth similarity loss $\mathcal{L}_d$ promotes $G$ to synthesize 3D geometries that are close to their corresponding source-domain geometries. As the source-domain geometries have valid geometric structures and correctly reflect the camera poses, $\mathcal{L}_d$ guides $G$ to synthesize valid unflattened geometries that correctly reflect the camera poses during the entire adaptation process.
Finally, $\mathcal{L}_n$ helps avoid degenerate 3D structures with high-frequency artifacts.
Thanks to the pose loss and geometric priors, the 3D synthesis network can be adequately adapted without drifting to an improper solution, which also helps the adaptation of the pose-estimation network.

\subsection{Training Details}
We pretrain the 3D synthesis network $G$ and the pose-estimation network $P$ on the real portrait images of the FFHQ dataset~\cite{stylegan}.
We apply horizontal flip for data augmentation. 
We use the Adam optimizer~\cite{adam} with learning rates of 0.0001 and 0.00125 for optimizing $P$ and $G$, respectively. The learning rate for the discriminator $D$ is 0.00075. 
The 3D synthesis and pose-estimation networks $G$ and $P$ are alternatively trained within a mini-batch of 32 images. 
We freeze the first 10 layers of the discriminator $D$ for stable domain adaptation~\cite{freeze-d}.
We use the weights $\alpha$, $\beta$ and $\gamma$ differently for target drawing domains as provided in the Supplemental Document.

\begin{figure*}[!t]
\centering
\includegraphics[width=0.98\linewidth]{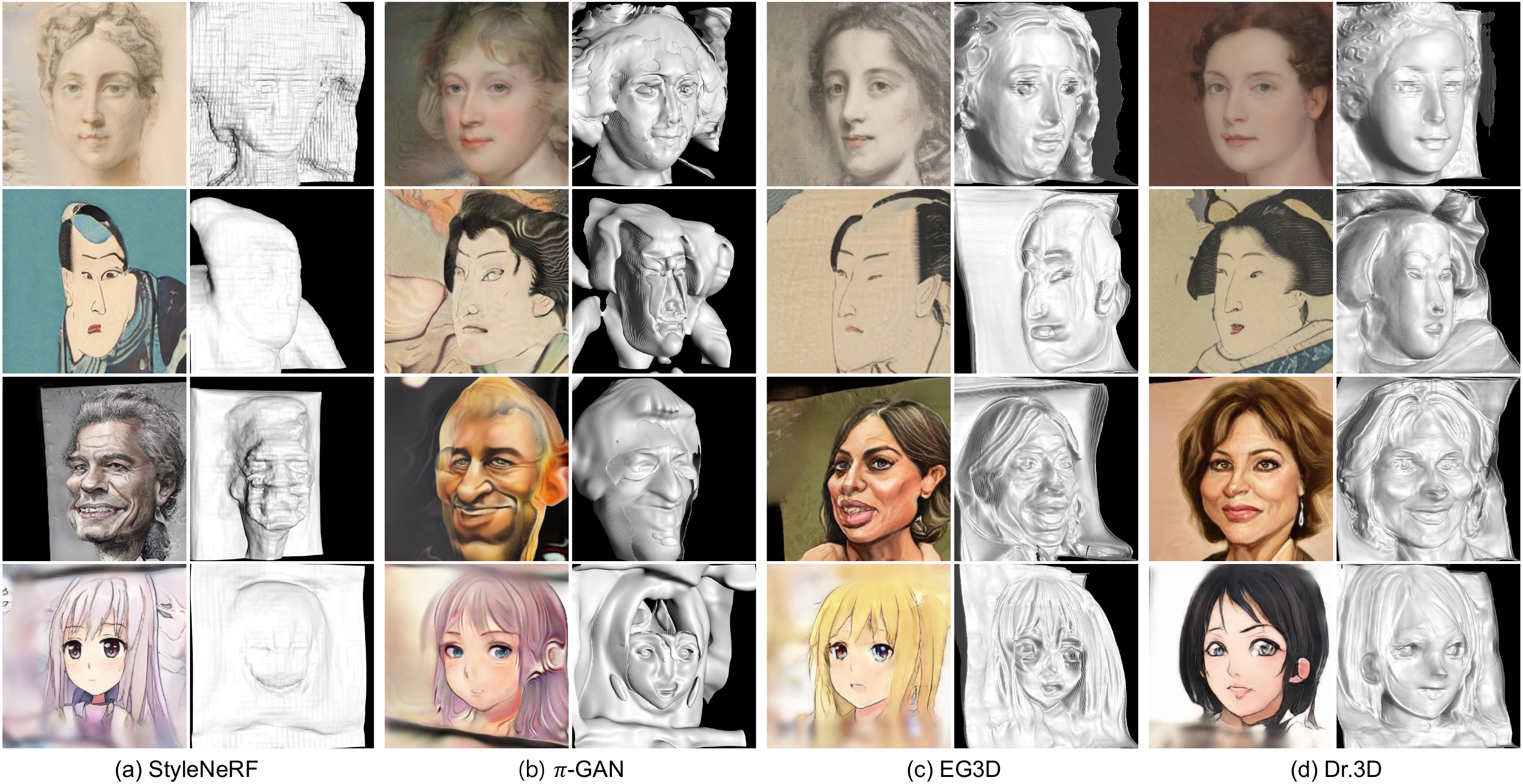}
\caption{Qualitative comparison among StyleNeRF~\cite{stylenerf}, $\pi$-GAN~\cite{pigan}, EG3D~\cite{eg3d} and ours.
The contents of the images are different as they are generated by differently trained generator models.
StyleNeRF, $\pi$-GAN and EG3D produce corrupted 3D geometries and unnatural-looking images especially for challenging styles such as ukiyo-e and anime, while our method produces more plausible shapes and images.
}
\label{fig:qualitative_comparison}
\end{figure*}

\section{Assessment}
\label{sec:results}
We conduct extensive validation of our method on four datasets of different drawing styles: historical art~\cite{stylegan-ada}, ukiyo-e~\cite{ukiyoe}, anime~\cite{anime}, and caricature~\cite{webcaricature}.
For the anime dataset, we crop and align face regions using an off-the-shelf face detection method~\cite{face_detection}.
We apply \MethodName{} to each dataset and obtain an adapted 3D GAN model separately.
Figure~\ref{fig:curated_examples} shows curated examples of 3D-aware drawing synthesis for the different drawing styles, demonstrating our 3D-aware synthesis capability for diverse drawing styles.
Refer to the Supplemental Document for uncurated results.

\subsection{Comparison}
We compare \MethodName{} to recent GAN-based 3D synthesis methods: StyleNeRF~\cite{stylenerf}, $\pi$-GAN~\cite{pigan} and EG3D~\cite{eg3d}. In the case of EG3D, we directly adapt EG3D from real photos to artistic drawings using camera poses estimated by an off-the-shelf pose-estimation network ~\cite{DECA}.
For the results of parametric fitting~\cite{DECA} and physics-based decomposition methods~\cite{unsup3d,gan2shape}, refer to the Supplemental Document.
Figure~\ref{fig:qualitative_comparison} shows a qualitative comparison between previous 3D GANs and ours.
Na\"{i}ve domain adaptation of the previous 3D GANs fails to handle diverse drawing shapes and appearances, resulting in low-fidelity images and flattened geometries.
\MethodName{} reconstructs plausible shapes and images of drawing, outperforming the previous methods.

We further conduct quantitative analysis on the fidelity of synthesized images and shapes.
The qualities of synthesized images are evaluated using FID~\cite{fid} and KID~\cite{kid}.
We use $256\times256$-sized images for all the methods except $\pi$-GAN, for which we use $128\times128$-sized images due to its large memory requirement.
Table~\ref{table:quantitative_img} shows the evaluation results where \MethodName{} achieves the best image-synthesis fidelity except for caricatures, thanks to our effective adaptation scheme.

Quantitative evaluation of synthesized shapes mandates the ground-truth shapes of drawings, which are challenging to obtain in most cases. 
For the historical-art dataset, as done in EG3D \cite{eg3d}, we obtain the \emph{pseudo} ground-truth shapes and poses of randomly generated drawings using a parametric fitting method \cite{DECA}. We measure depth and pose error by calculating MSE between generated sets and pseudo ground-truth depths and poses.
For the evaluation of caricatures, we utilize the 3DCaricShop dataset, which provides paired images and 3D shapes created by artists. We reconstruct caricature images using GAN-inversion and measure depth and pose error with ground-truth geometries.

Table~\ref{table:quantitative_shape} shows that \MethodName{} generally performs better than the other methods in terms of shapes and poses. 
While StyleNeRF achieves better depth accuracy than ours for the historical-art dataset, it shows the worst pose accuracy.
Also, while the table shows that EG3D achieves comparable results to ours, it tends to produce noisy and flattened shapes as shown in Figure~\ref{fig:qualitative_comparison}.

\begin{table}[!t]
\caption{Quantitative comparison on the image quality among $\pi$-GAN~\cite{pigan}, StyleNeRF~\cite{stylenerf}, EG3D~\cite{eg3d} and ours.}
\vspace{-3mm}
\begin{tabular}{@{}c|c|cccc@{}}
\hline
          &                  & Hist.~art          & Ukiyo-e           & Anime             & Caricature           \\ \hline
$\pi$-GAN & FID $\downarrow$ & 46.40             & 65.91             & 48.78             & 73.25                 \\
          & KID $\times 10^3 \downarrow$ & 26.14             & 53.79             & 28.29             & 52.15                 \\ \hline
StyleNeRF & FID $\downarrow$ & 34.99             & 58.52             & 27.94             & 22.53                 \\
          & KID $\times 10^3\downarrow$ & 14.51             & 58.72             & 12.41             & 11.72                 \\ \hline
EG3D      & FID $\downarrow$ & 26.95             & 40.16             & 20.75             & \textbf{15.71}        \\
          & KID $\times 10^3\downarrow$ & 9.295             & 32.94             & 8.699             & \textbf{7.123}        \\ \hline
\MethodName{}      & FID $\downarrow$ & \textbf{23.42}    & \textbf{37.38}    & \textbf{18.74}    & 19.69        \\
(Ours)          & KID $\times 10^3\downarrow$ & \textbf{5.916}    & \textbf{29.65}    & \textbf{6.335}    & 9.180        
                \\ \hline
\end{tabular}
\label{table:quantitative_img}
\end{table}

\begin{table}[!t]
\caption{
Quantitative comparison on the shape and pose quality among $\pi$-GAN~\cite{pigan}, StyleNeRF~\cite{stylenerf}, EG3D~\cite{eg3d} and \MethodName{}. 
}
\begin{tabular}{@{}c|cc|cc@{}}
\hline
          & \multicolumn{2}{c|}{Hist.~art} & \multicolumn{2}{c}{Caricature} \\ 
          & Depth         & Pose          & Depth          & Pose          \\ \hline
$\pi$-GAN & 0.305         & 0.072         & 0.151          & 0.077                    \\
StyleNeRF & \textbf{0.169}& 0.333         & 0.688          & 0.326                    \\
EG3D      & 0.215         & 0.054         & 0.033          & \textbf{0.047}     \\
\MethodName{} (Ours)      & 0.217         & \textbf{0.030}& \textbf{0.020} & 0.070 \\ \hline
\end{tabular}
\label{table:quantitative_shape}
\end{table}

\begin{figure}[t]
\centering
\includegraphics[width=\linewidth]{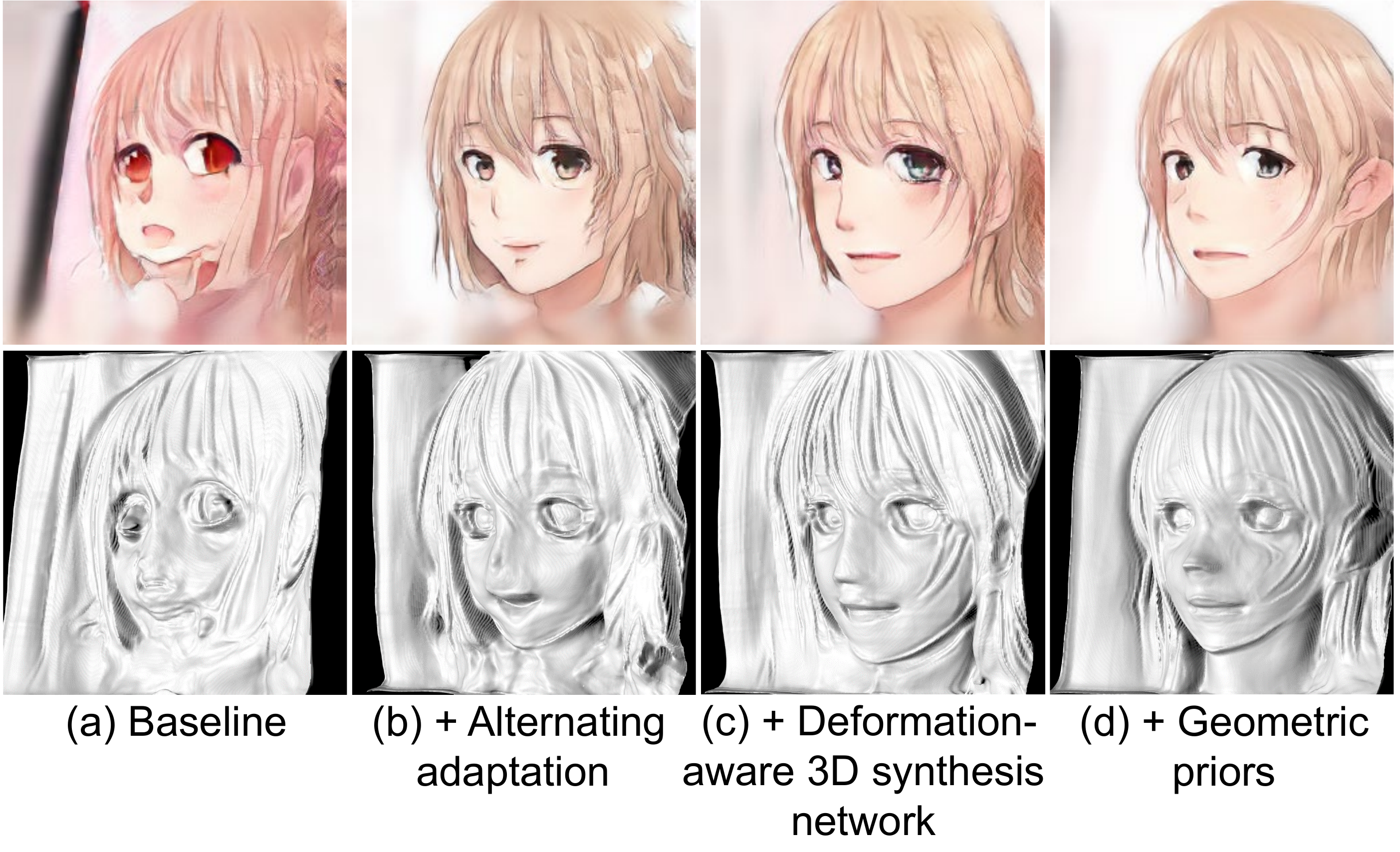}
\vspace{-0.5cm}
\caption{Ablation study. 
The baseline model (EG3D~\cite{eg3d}) synthesizes a distorted image and a flattened geometry as shown in (a).
While our alternating adaptation helps avoid flattened shapes as shown in (b),
our deformation-aware 3D synthesis network, and geometric priors further improve the synthesis quality.
}
\label{fig:ablation_img}
\end{figure}

\begin{figure}[t]
\centering
\includegraphics[width=0.9\linewidth]{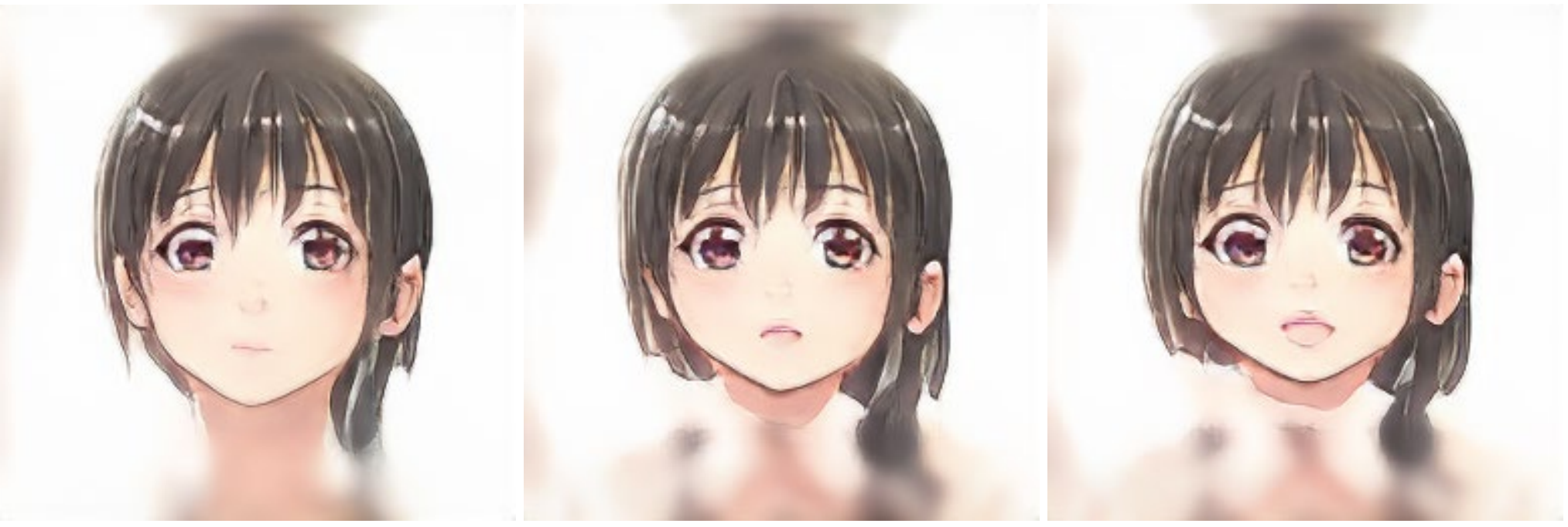}
\vspace{-0.2cm}
\caption{
Image synthesis results from different deformation codes $z_d$. For all the results, the same latent code $z$ is used.
}
\label{fig:deformation_img}
\end{figure}

\begin{figure}[t]
\centering
\includegraphics[width=\linewidth]{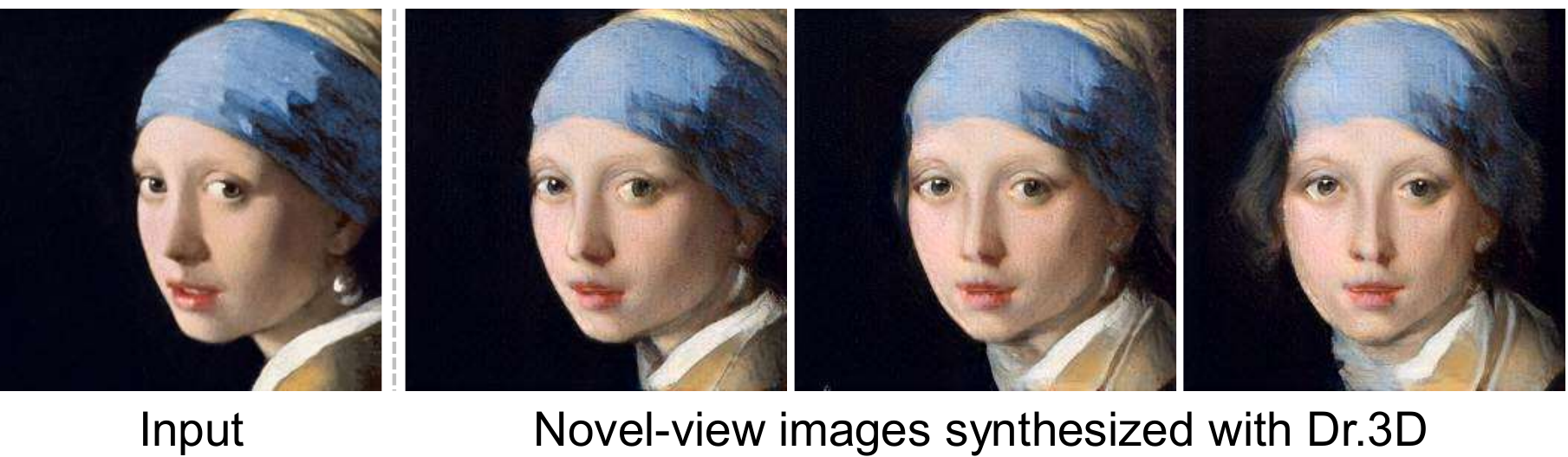}
\vspace{-0.2cm}
\caption{Novel view synthesis of a real-world drawing.
Input: Girl with a Pearl Earring, 1665 by Johannes Vermeer, WikiArt [Public Domain] via (https://bit.ly/3PE66CT).
}
\label{fig:synthesis_img}
\end{figure}

\subsection{Ablation Study}
\MethodName{} effectively deals with the intrinsic ambiguity of drawing images by means of (1) a deformation-aware 3D synthesis network, (2) alternating adaptation of pose estimation and image synthesis, and (3) geometric priors. 
We assess the impact of each component by starting with our baseline network, EG3D~\cite{eg3d}. 
Figure~\ref{fig:ablation_img} shows an ablation result. 
Using the original EG3D model on drawings results in a flattened shape (Figure~\ref{fig:ablation_img}(a)).   
For training the EG3D model, we used the camera pose estimated from an off-the-shelf pose-estimation network~\cite{DECA}. 
Our alternating adaptation of the pose-estimation network and the deformation-aware 3D synthesis network enables us to recover a better 3D geometry (Figure~\ref{fig:ablation_img}(b)). 
Adding our deformation-aware 3D synthesis network further improves the shape-reconstruction fidelity and the quality of synthesized images as it helps capture shape and style variations in drawings (Figure~\ref{fig:ablation_img}(c)).
Our full method, \MethodName{}, with the geometric priors, results in the best synthesis quality for both image and shape (Figure~\ref{fig:ablation_img}(d)).

As discussed in Section~\ref{sec:deform_net}, drawings have a larger distribution of potentially-feasible 3D shapes than photos.
Our deformation network helps model such a larger distribution of drawings by expanding the representation space with an additional latent code $z_d$, which leads to higher-quality adaptation results as shown in Figure~\ref{fig:ablation_img}(c).
Figure~\ref{fig:deformation_img} shows another example of the deformation network.
In the figure, while the same latent code $z$ synthesizes all the images, they exhibit different details due to different deformation codes $z_d$, proving the larger representation space expanded by the deformation network.
More analysis on the deformation-aware network is provided in the Supplemental Document.

\begin{figure}[t]
\centering
\includegraphics[width=0.95\linewidth]{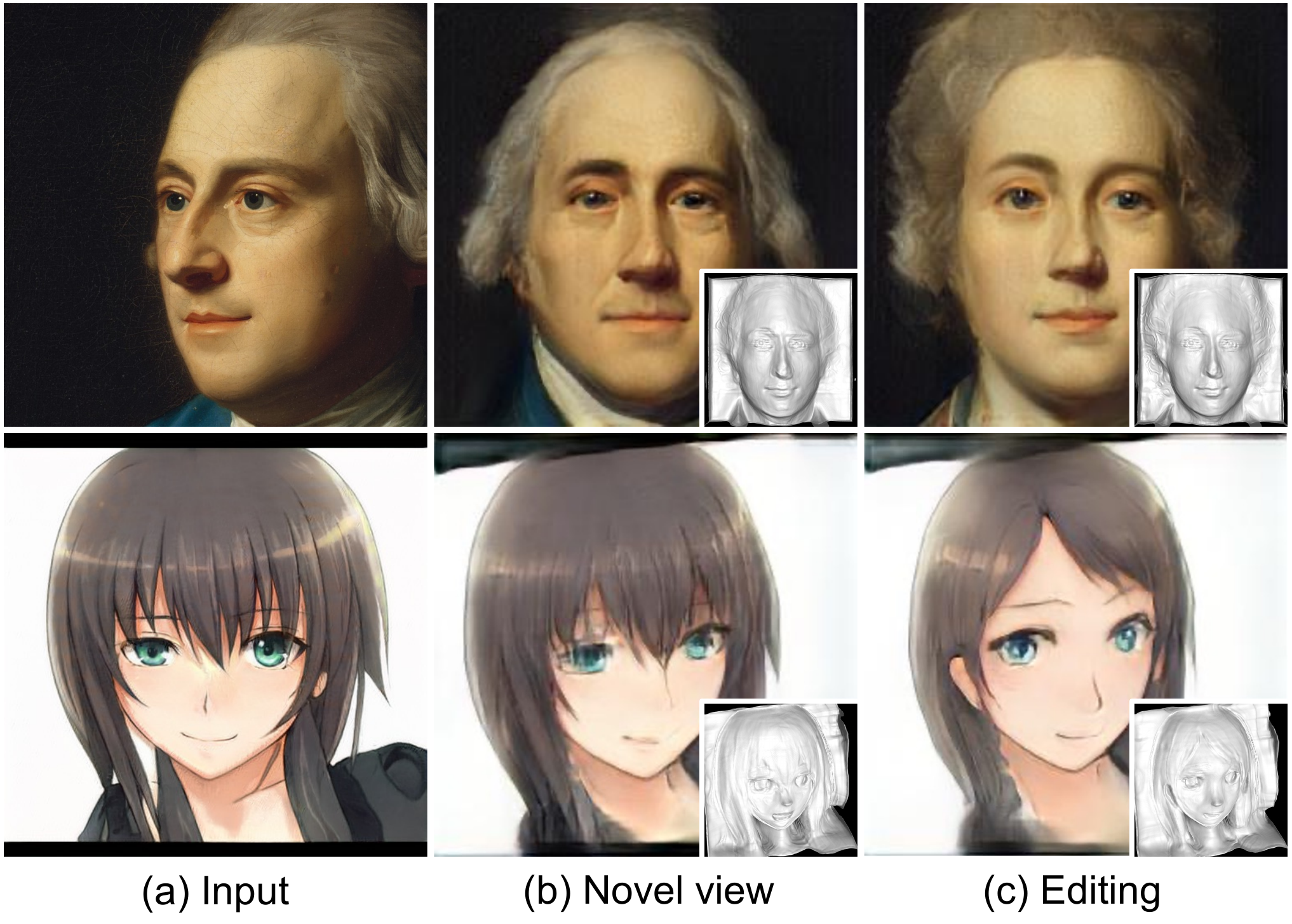}
\vspace{-0.2cm}
\caption{
Semantic editing of input drawings.
Top: male to female.
Bottom: hairstyle change.
Input on the top row: Gulian Verplanck, 1771 by John Singleton Copley, WikiArt [Public Domain] via (https://bit.ly/3PE6vVV).
}
\label{fig:editing}
\end{figure}

\subsection{3D-aware Semantic Editing of Drawing}
Combining GAN inversion with \MethodName{} enables multi-view consistent editing of real-world drawings such as novel-view synthesis and semantic editing.
Figure~\ref{fig:synthesis_img} shows examples of novel-view synthesis of real-world drawings.
In these examples, we estimate the camera poses of the input images using our domain-adapted pose-estimation network and invert the images to GAN latent codes using the pivotal tuning inversion method~\cite{pti}.
Then, we synthesize novel views of the input images by feeding their latent codes and new camera poses to the 3D synthesis network.

\MethodName{} also enables multi-view consistent semantic editing on real-world drawings.
Figure~\ref{fig:editing} shows examples of semantic editing.
In these examples, we use editing vectors found by applying InterfaceGAN~\cite{interfacegan} using the original EG3D network trained on the FFHQ dataset.
\section{Conclusion}
This paper presented \MethodName{}, a novel 3D GAN adaptation method from real portraits to artistic drawings. 
To handle the intrinsic geometric ambiguity of drawings, we proposed alternating adaptation of the pose estimation and image synthesis, a deformation-aware network, and geometric priors.
We experimentally validated that our approach can successfully adapt 3D GANs to drawings for the first time.
\MethodName{} allows to edit an artistic drawing in consideration of its 3D geometric structure and semantics of the content. 

\paragraph{Limitations} 
While \MethodName{} can produce superior results to previous methods,
it may still produce flattened geometries for some latent codes for challenging domains such as anime.
Refer to the Supplemental Document for a failure example.
Also, our method is limited in dealing with the background region in which 3D-consistent shared geometric features do not exist in training images.
We note that this limitation also applies to existing 3D-GAN methods including EG3D~\cite{eg3d}.
One potential way to resolving this would be to divide the feature-generation procedure into two: one for the foreground and the other for the background~\cite{stylenerf}.
Extending \MethodName{} to diverse target domains including non-human faces would also be an interesting future direction.

\begin{acks}
This research was supported by \grantsponsor{IITP}{IITP}{} grants funded by the Korea government (MSIT) (\grantnum[]{IITP}{2021-0-02068}, \grantnum[]{IITP}{2019-0-01906}), an \grantsponsor{NRF}{NRF}{} grant funded by the the Korea government (MOE) (\grantnum[]{NRF}{2022R1A6A1A03052954}), and \grantsponsor{Pebblous}{Pebblous}{}.
\end{acks}

\nocite{*}
\bibliographystyle{ACM-Reference-Format}
\bibliography{Dr.3D}


\end{document}